\documentclass[letterpaper]{article} 
\usepackage[preprint]{aaai2027}  
\usepackage[hyphens]{url}  
\usepackage{graphicx} 
\usepackage{cuted}
\usepackage{natbib}  
\usepackage{caption} 
\usepackage{algorithm}
\usepackage{algorithmic}

\usepackage{newfloat}
\usepackage{listings}
\DeclareCaptionStyle{ruled}{labelfont=normalfont,labelsep=colon,strut=off} 
\floatstyle{ruled}
\newfloat{listing}{tb}{lst}{}
\floatname{listing}{Listing}

\usepackage{booktabs}

\title{MATE: Multi-Agent Virtual Teleoperation Platform \\for Humanoid Collaboration Data Collection}
\author{
    Yichuan Yu\textsuperscript{\rm 1},
    Youzhuo Wang\textsuperscript{\rm 1},
    Yiming Ren\textsuperscript{\rm 1,2},
    Di Feng\textsuperscript{\rm 1},
    Yexuan Yang\textsuperscript{\rm 1},
    Bingxi Yang\textsuperscript{\rm 1},
    Shengxiao Gong\textsuperscript{\rm 1},
    Yujing Sun\textsuperscript{\rm 2}\corresponding,
    Yuexin Ma\textsuperscript{\rm 1}\corresponding
}
\affiliations{
    \textsuperscript{\rm 1}ShanghaiTech University\\
    \textsuperscript{\rm 2}Nanyang Technological University
}

\begin{document}

\maketitle

\begin{strip}
    \centering
    \includegraphics[width=1.0\textwidth]{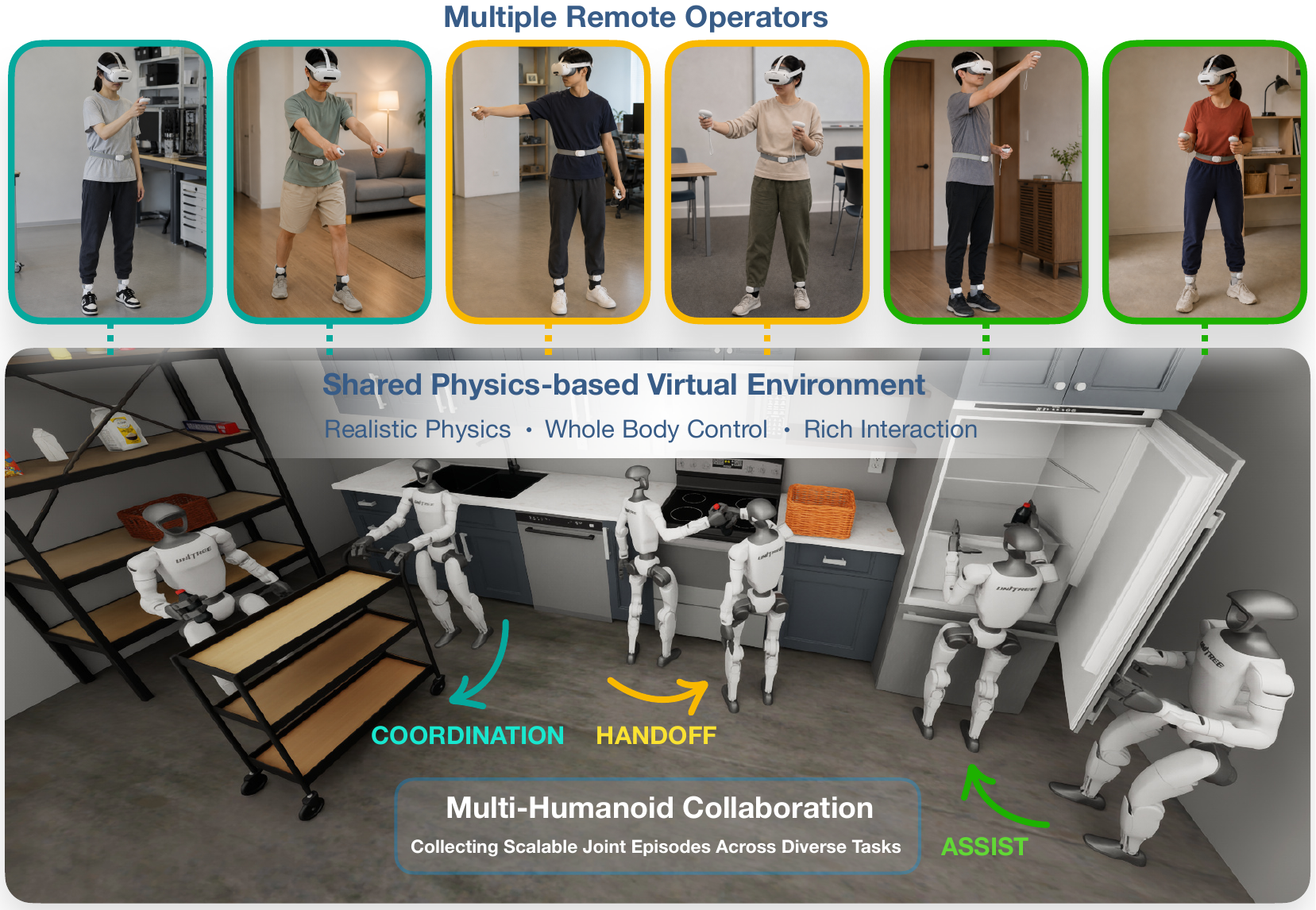}
    \captionof{figure}{Conceptual overview of MATE, a multi-agent virtual teleoperation platform for scalable humanoid collaboration data collection. Multiple remote operators simultaneously control whole-body humanoids in a shared physics-based virtual environment, enabling the collection of physically coupled collaboration episodes across diverse long-horizon tasks.}
    \label{fig:teaser}
\end{strip}

\begin{abstract}
Humanoid robots require diverse embodied experiences to acquire complex loco-manipulation and collaborative skills. However, existing humanoid data pipelines primarily focus on individual agents, while physical multi-robot collaboration remains difficult to scale due to costly hardware, dedicated spaces, and repeated resets. In this work, we introduce \textbf{MATE}, a \textbf{M}ulti-\textbf{A}gent virtual \textbf{TE}leoperation platform for humanoid collaboration data collection that enables multiple geographically distributed operators to simultaneously control whole-body humanoids in a shared physics-based environment. \textbf{MATE} removes the need for multiple physical robots and co-located operation while preserving physically coupled interactions among humanoids, objects, and environments.
Using \textbf{MATE}, we construct a multi-humanoid collaboration dataset comprising 24.1 hours of coordinated behavior across 2,500 joint episodes and five long-horizon tasks, including object handover, relay delivery, environment interaction, and cooperative transport. To improve learning from these interaction-rich demonstrations, we introduce \textbf{EAIS}, an \textbf{E}xecution-\textbf{A}ligned \textbf{I}nteraction \textbf{S}ampling strategy that computes sampling signals within an execution-aligned prefix and prioritizes task-progressing and interaction-critical behaviors.
We evaluate \textbf{MATE} with representative imitation learning and vision-language-action policies across diverse collaboration tasks. Experiments demonstrate efficient data collection, effective policy learning, and zero-shot transfer from virtual demonstrations to a physical humanoid without real-world fine-tuning.
\textbf{Project page:} \url{https://yerik-yu.github.io/MATE}
\end{abstract}
\section{Introduction}
Humanoid robots are emerging as a promising embodiment for artificial intelligence, with the potential to operate in complex human-centric environments and assist with diverse physical tasks. Beyond individual loco-manipulation abilities, future humanoid systems will need to perceive, coordinate, and collaborate with other embodied agents in settings where successful collaboration depends on reasoning about shared environments, dynamic interactions, and long-horizon task dependencies. Achieving such capabilities requires large-scale and diverse embodied experiences that capture not only individual robot skills but also the physical and temporal dependencies underlying collaboration.

Current humanoid data collection approaches have made significant progress through real-world teleoperation~\citep{seo2023trill,he2024omnih2o,ze2025twist2} and simulation-based generation~\citep{he2025viral,yu2026oasis}. However, scaling these pipelines to multi-humanoid collaboration remains challenging. In physical environments, collaborative data collection requires multiple capable robots, larger workspaces, synchronized operators, and frequent coordination for initialization and recovery, resulting in substantial hardware and operational costs. In simulation, simply duplicating single-agent pipelines is insufficient, as collaborative behaviors depend on the coupled evolution of multiple agents, shared objects, and environments, requiring synchronized control, consistent physics interaction, and coordinated data recording. Consequently, existing humanoid data collection systems largely remain centered on individual agents and lack a scalable paradigm for acquiring long-horizon collaborative experiences.

Recent works also explore virtual reality (VR) teleoperation in simulation as a promising direction that combines human-provided demonstrations with the scalability of virtual environments. For example, SIMPLE~\citep{wei2026simple} integrates VR-based human demonstration with physics simulation and photorealistic rendering, enabling efficient humanoid loco-manipulation data collection and successful sim-to-real transfer. However, these approaches still primarily follow a single-agent paradigm, where one operator controls one humanoid in isolation. Extending VR-based humanoid teleoperation toward multi-humanoid collaboration requires preserving not only individual robot skills but also interaction dependencies among agents, objects, and environments. The behavior of one humanoid can continuously alter the observations and feasible actions of others, making scalable collection of long-horizon collaborative experiences an open challenge.

In this work, we introduce MATE, a virtual reality-based platform that enables scalable acquisition of multi-humanoid collaboration demonstrations. As illustrated in Fig.~\ref{fig:teaser}, MATE assigns each operator to an individual humanoid through independent embodied interfaces and executes all agents within a shared physics-based environment. Unlike independently collected single-robot trajectories, MATE synchronizes multi-agent control streams and records physically coupled joint episodes, where the actions of one humanoid dynamically influence the observations, object states, and feasible behaviors of other agents. This design enables the collection of long-horizon collaboration behaviors, including handover, coordination, and cooperative manipulation, without requiring multiple physical robots or co-located operation.
However, the resulting collaboration demonstrations introduce new challenges for policy learning. Long-horizon multi-agent trajectories contain highly imbalanced temporal structures: frequent locomotion, approach, and waiting behaviors occupy most of the demonstrations, while sparse interaction-critical transitions such as contact establishment, release, and role changes provide essential supervision. To address this challenge, we introduce EAIS, an execution-aligned interaction sampling strategy that computes sampling signals within the execution-aligned prefix of each demonstration window and prioritizes segments with sustained task progress and interaction significance.

We evaluate MATE as a data collection paradigm for humanoid collaboration through extensive experiments on long-horizon loco-manipulation tasks. We benchmark representative imitation learning and vision-language-action policies trained on MATE demonstrations, with one humanoid controlled by the learned policy and the other following partner trajectories that vary in motion and timing across rollouts. We also analyze the scalability advantage of virtual multi-agent teleoperation compared with physical collection and evaluate EAIS under identical training conditions. Finally, we demonstrate the transferability of MATE-generated demonstrations through a zero-shot sim-to-real deployment on a physical humanoid without real-world fine-tuning. These results show that MATE provides a scalable and effective pathway for acquiring and utilizing interaction-rich humanoid collaboration experiences.
Our contributions are summarized as follows:

\begin{enumerate}
\item We introduce MATE, a multi-agent virtual teleoperation platform that enables scalable collection of physically coupled humanoid collaboration demonstrations in shared virtual environments.
\item We construct a multi-humanoid collaboration dataset consisting of long-horizon joint episodes covering diverse coordination and loco-manipulation behaviors.
\item We introduce EAIS, an execution-aligned interaction sampling strategy for learning from long-horizon collaborative demonstrations, and validate MATE through multi-policy evaluation and zero-shot sim-to-real transfer.
\end{enumerate}

To support reproducibility and future research, we will release the MATE platform, dataset, task configurations, and EAIS implementation upon publication.

\section{Related Work}

\subsection{Humanoid Data Collection}

Teleoperation provides an effective way to acquire expert robot demonstrations, but real-world collection remains constrained by hardware cost, operator effort, and repeated environment resets. Recent systems such as ALOHA and Open-TeleVision improve the accessibility of robot data collection through low-cost and immersive interfaces~\citep{zhao2023aloha,cheng2024opentelevision}. 

For humanoid robots, whole-body teleoperation requires coordinated human-to-robot motion mapping, locomotion, and manipulation. Recent systems advance whole-body teleoperation and demonstration collection~\citep{he2024h2o,fu2024humanplus,seo2023trill,he2024omnih2o,ben2025homie,ze2025twist,ze2025twist2}, while SONIC provides scalable whole-body motion tracking and control~\citep{luo2025sonic}. Humanoid Everyday expands the scale and diversity of real-world humanoid data, while HuMI explores efficient robot-free collection of whole-body demonstrations~\citep{zhao2025humanoideveryday,nai2026humi}.
Simulation-based approaches provide an alternative solution by reducing hardware dependency and enabling scalable scene generation and trajectory collection. AgentWorld, VIRAL, SIMPLE, and OASIS demonstrate the potential of simulation for large-scale humanoid learning and sim-to-real transfer \citep{zhang2025agentworld,he2025viral,wei2026simple,yu2026oasis}. However, existing data collection paradigms mainly focus on individual robot behaviors, while scalable acquisition of multi-agent embodied experiences remains underexplored.

\subsection{Multi-Agent Collaboration}

Collaborative behaviors require robots to reason about the actions, states, and intentions of other agents. Existing studies have explored multi-agent coordination through decentralized reinforcement learning~\citep{lowe2017maddpg,yu2022mappo}, language-based task planning~\citep{mandi2023roco}, and embodied robot collaboration, including coordinated manipulation and decentralized multi-embodiment control~\citep{song2025collabot,doshi2026chorus}. CooHOI and SynAgent extend single-agent motion or interaction priors to cooperative object manipulation with multiple humanoid characters~\citep{gao2024coohoi,yao2026synagent}. Together, these works demonstrate diverse approaches to collaborative policy learning across humanoid characters and robotic embodiments.

Despite these advances in collaborative behavior and policy learning, collecting synchronized, physically coupled multi-agent interaction data remains challenging. Such data preserves the temporal and physical dependencies through which one agent's actions alter shared object and contact states, partner observations, and subsequent feasible behaviors; these dependencies are not directly captured by independently collected single-agent demonstrations. Recent works investigate multi-agent imitation learning from single-agent demonstrations~\citep{mattson2025r2bc} and synchronized dual-robot teleoperation~\citep{zhao2026duet}. However, physical multi-robot data collection remains difficult to scale due to robot cost, workspace requirements, and repeated resets. MATE complements these efforts by enabling scalable synchronized multi-humanoid demonstration collection through multi-user virtual teleoperation.

\subsection{Long-Horizon Robot Learning}
Long-horizon robot tasks require policies to capture complex temporal dependencies across locomotion, manipulation, and interaction. Recent imitation learning approaches, including action chunking~\citep{zhao2023aloha}, diffusion-based policies~\citep{chi2023diffusionpolicy}, and vision-language-action models~\citep{wei2026psi0,nvidia2026grootn17,physicalintelligence2025pi05}, have improved the modeling of long-duration behaviors from demonstrations.
However, long-horizon trajectories often contain substantial temporal imbalance, where frequent motions such as navigation and waiting dominate the data distribution while critical events, including contact, handover, and coordination transitions, occur sparsely. Existing temporal selection and decomposition methods mainly rely on signals derived from individual-agent trajectories~\citep{zhang2023uvd,kou2024kisa,mark2026bpp} and do not explicitly account for interaction-dependent behaviors. In multi-agent humanoid collaboration, low-motion states may still encode meaningful coordination, motivating our execution-aligned interaction sampling strategy for learning from collaborative demonstrations.

\section{Methodology}

\begin{figure*}[ht!]
    \centering
    \includegraphics[width=1.0\linewidth]{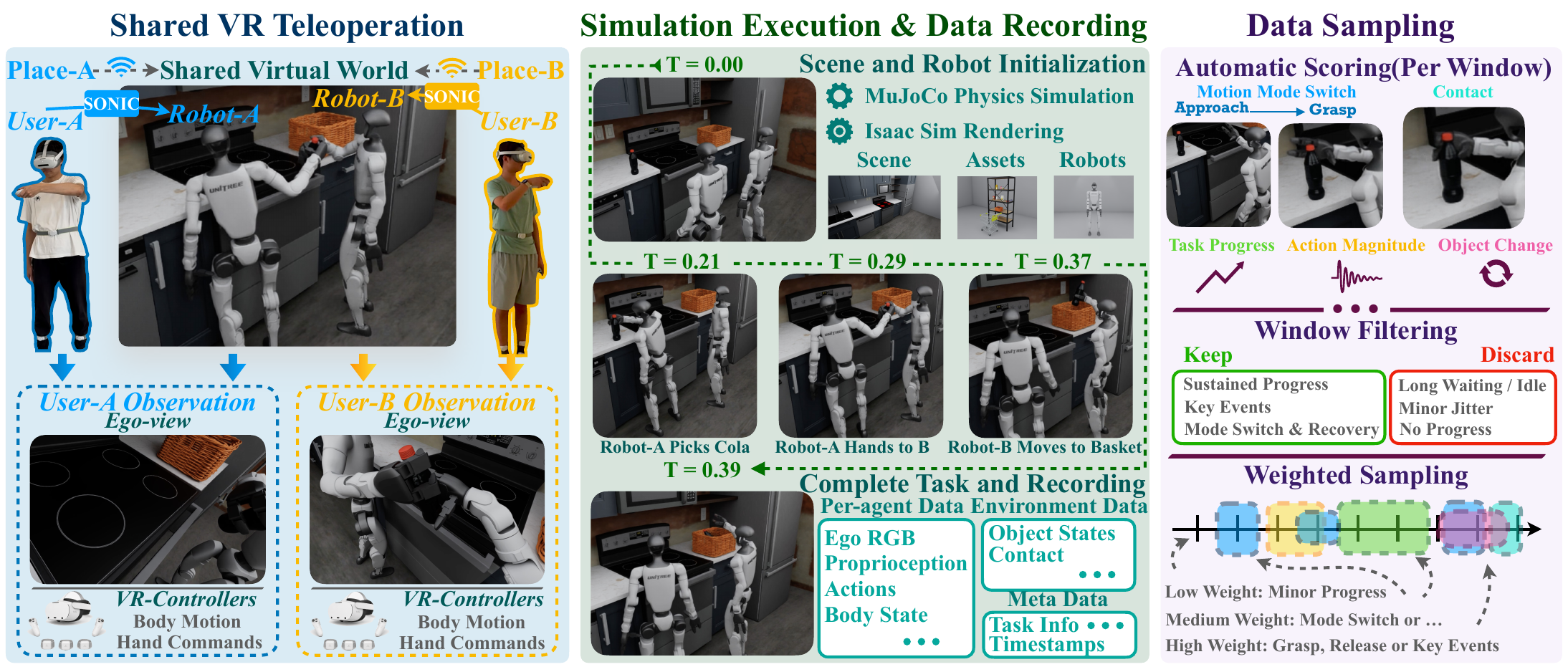}
    \caption{MATE pipeline for multi-user virtual teleoperation and execution-aligned interaction sampling.}
    \label{fig:pipeline}
\end{figure*}
MATE addresses two challenges in humanoid collaboration learning: scalable acquisition of physically coupled demonstrations and effective utilization of long-horizon interaction trajectories. As illustrated in Fig.~\ref{fig:pipeline}, MATE first collects synchronized, physically coupled joint episodes in a shared simulation and then applies EAIS to prioritize execution-relevant training windows.
During data collection, each operator controls an individual whole-body humanoid through an independent VR interface. The synchronized simulation execution preserves the interaction dynamics among humanoids, objects, and environments, producing joint episodes that capture collaborative behaviors beyond isolated robot trajectories. These episodes provide the foundation for learning long-horizon humanoid collaboration.
However, collaboration demonstrations contain highly imbalanced temporal structures, where frequent motions may dominate training while sparse interaction-critical behaviors remain underrepresented. To address this challenge, we introduce \textbf{Execution-Aligned Interaction Sampling (EAIS)}, which computes sampling signals within an execution-aligned prefix and prioritizes segments containing task progress and interaction events.

\subsection{MATE Platform}

MATE enables scalable collection of humanoid collaboration demonstrations through a multi-user virtual teleoperation system. Each operator is associated with a designated humanoid agent and interacts with the robot through an independent VR-based embodied interface. The operator receives robot-centric visual feedback and generates whole-body commands through the SONIC control interface~\citep{luo2025sonic}, including locomotion, body motion, and hand articulation. This one-to-one operator-humanoid mapping allows multiple users to simultaneously control different humanoids while maintaining independent command streams.

During teleoperation, control commands from all active humanoid agents are transmitted through independent communication channels and synchronized at each simulation step. The synchronized commands are jointly applied to a shared MuJoCo physics environment~\citep{todorov2012mujoco}, where contacts, object dynamics, and articulated interactions are resolved consistently. Unlike independently collected single-agent trajectories, MATE therefore preserves the causal coupling between agents: an action performed by one humanoid can immediately change the object states, observations, and feasible actions of other humanoids.

The shared simulation execution is coupled with a high-fidelity rendering pipeline for visual data collection. While MuJoCo is responsible for efficient physics simulation, Isaac Sim~\citep{gao2026isaacsim} serves as the rendering backend to generate robot-centric visual observations. This separation enables scalable simulation execution while providing configurable visual appearances and camera viewpoints for downstream learning and sim-to-real evaluation.

Each execution is recorded as a joint collaboration episode under a shared simulation clock. For an episode with $N$ humanoid agents and horizon $T$, MATE records

\begin{equation}
\tau =
\left\{
\left(o_t^i,a_t^i\right)_{i=1}^{N},
s_t^{env}
\right\}_{t=1}^{T},
\end{equation}
where $o_t^i$ and $a_t^i$ denote the observation and action of agent $i$, and $s_t^{env}$ represents the shared environment state, including object configurations and interaction information. By preserving synchronized multi-agent trajectories and environment evolution, MATE produces physically coupled joint episodes rather than isolated robot demonstrations.

\subsection{Execution-Aligned Interaction Sampling}

Although MATE provides physically coupled humanoid collaboration episodes, directly training on uniformly sampled windows can introduce temporal supervision imbalance. Human teleoperation demonstrations often contain redundant behaviors, such as hesitation, unnecessary adjustment, prolonged waiting, and minor motion fluctuations. These segments may occupy a large fraction of long-horizon trajectories, while sparse but critical behaviors, including grasping, release, contact establishment, and coordination transitions, provide essential supervision for policy learning. Therefore, EAIS aims to construct a training distribution that emphasizes behaviors contributing to successful task execution. 

Given a demonstration episode $\tau$, a candidate training window for agent $i$ starting at timestamp $t$ is defined as

\begin{equation}
W_t^i=(o_t^i,a_{t:t+H-1}^i),
\end{equation}
where $i$ indexes the robot whose training windows are being evaluated, $o_t^i$ denotes its observation at time $t$, $a_{t:t+H-1}^i$ denotes its demonstrated action sequence, and $H$ is the prediction horizon. EAIS evaluates each window over an execution-aligned prefix of length $H_e \leq H$ and computes the sampling signals from the corresponding synchronized interaction trajectory within this prefix.

For each candidate window, EAIS performs automatic scoring using label-free signals available from the recorded interaction trajectory. Specifically, the scoring process considers five complementary signals: (1) task progress measured by embodiment and object state changes, (2) motion-mode transitions indicating changes between locomotion and manipulation behaviors, (3) action magnitude changes capturing behavioral transitions, (4) object state changes reflecting task consequences, and (5) interaction events including contact establishment and release. Task progress is evaluated over different embodiment components. For a state group $g$, the observable displacement is computed as

\begin{equation}
D_g^i(t)=
\sqrt{
\frac{1}{|g|}
\sum_{j\in g}
\left\|
x_{t+H_e,j}-x_{t,j}
\right\|^2
},
\end{equation}
where $g$ denotes a group of state variables for robot $i$, including the robot base, body joints, hands, and manipulated objects; $|g|$ is the number of variables in the group; $j$ indexes an individual state variable; and $x_{t,j}$ represents its value at time $t$. These signals are combined to identify windows containing sustained task progress, key interaction events, and meaningful behavior transitions.
After automatic scoring, EAIS assigns different sampling priorities to candidate windows. Windows dominated by prolonged waiting, idle behavior, or minor jitter without corresponding object-state changes, interaction events, or measurable task progress are down-weighted or assigned zero sampling probability. In contrast, windows containing sustained task progress, coordination-relevant waiting, grasping, release, contact transitions, or other interaction-significant behavior changes receive higher priority. The final sampling weight is defined as

\begin{equation}
w_t^i=w_t^{i,\mathrm{prog}}\cdot w_t^{i,\mathrm{int}},
\end{equation}
where $w_t^{i,\mathrm{prog}}$ represents the task progression score and $w_t^{i,\mathrm{int}}$ represents interaction importance. The resulting training distribution preserves long-range behavioral context while increasing the representation of collaboration-critical segments, without requiring manually annotated task phases. \textbf{The detailed parameters and event-weighting rules are provided in the supplementary material.}
\begin{table*}[t!]
\centering
\footnotesize
\setlength{\tabcolsep}{4pt}
\renewcommand{\arraystretch}{1.08}

\begin{tabular*}{\textwidth}{@{\extracolsep{\fill}}lcccccc@{}}
\toprule
\textbf{Approach}
&
\textbf{Interaction}
&
\textbf{Embodiment}
&
\textbf{Data Type}
&
\textbf{Source}
&
\textbf{Scale}
&
\textbf{Tasks}
\\
\midrule

TRILL~\citeyearpar{seo2023trill}
& Single-agent
& Humanoid
& Trajectory
& Sim.+Real
& 2K traj.
& 8
\\

TWIST2~\citeyearpar{ze2025twist2}
& Single-agent
& Humanoid
& Trajectory
& Real
& 220 demos
& 2
\\

Humanoid Everyday~\citeyearpar{zhao2025humanoideveryday}
& Single-agent
& Humanoid
& Trajectory
& Real
& 10.3K traj.
& 260
\\

SIMPLE~\citeyearpar{wei2026simple}
& Single-agent
& Humanoid
& Trajectory
& Sim.
& 6K+ traj.
& 60
\\

OASIS~\citeyearpar{yu2026oasis}
& Single-agent
& Humanoid
& Trajectory
& Sim.
& 200 src. / 4K render.
& 4
\\

DUET~\citeyearpar{zhao2026duet}
& Multi-agent
& Heterogeneous robots
& Joint episode
& Real
& 286 eps.
& 4
\\

\textbf{MATE}
& \textbf{Multi-agent}
& \textbf{Humanoid pair}
& \textbf{Joint episode}
& \textbf{Sim.}
& \textbf{2.5K eps.}
& \textbf{5}
\\

\bottomrule
\end{tabular*}
\caption{
Comparison of representative humanoid demonstration datasets and collection frameworks.
}
\label{tab:dataset_comparison}
\end{table*}
\section{Dataset}
\label{sec:dataset}

Using MATE, we construct a multi-humanoid collaboration dataset consisting of long-horizon joint episodes collected through multi-user virtual teleoperation. As summarized in Table~\ref{tab:dataset_comparison}, existing humanoid datasets primarily focus on single-agent trajectories, covering individual locomotion or manipulation skills. In contrast, MATE targets multi-agent humanoid collaboration by collecting physically coupled joint episodes, where multiple humanoids interact within the same evolving environment. Compared with prior real-world or simulation-based datasets, MATE combines scalable virtual collection with embodied collaboration. Each episode preserves the coupled evolution of multiple humanoids, objects, and environments, enabling learning from interaction-rich behaviors rather than isolated robot trajectories.

\subsection{Collaborative Humanoid Episodes}
A key property of MATE is that the fundamental data unit is a joint episode rather than an independent robot trajectory. Each episode records the coordinated execution of two humanoids within a shared physical environment, preserving the interaction history among agents, objects, and task states. The dataset contains 500 joint episodes for each of five tasks, resulting in 2,500 collaborative episodes and 5,000 robot-centric trajectories. Demonstrations are collected at 50 Hz, yielding approximately 4.35 million time steps (24.1 hours) of coordinated humanoid behavior. Unlike independently collected single-agent demonstrations, a joint episode captures how one humanoid changes the future observations and feasible behaviors of another agent through shared object states and physical interactions. This property enables the study of coordination behaviors such as waiting, handover, cooperative manipulation, and role transitions.

\subsection{Long-Horizon Collaboration Tasks}
The MATE dataset contains five long-horizon humanoid collaboration tasks designed to cover representative interaction patterns. Each task requires a combination of locomotion, manipulation, temporal coordination, and physical interaction between humanoid agents.

\noindent\textbf{Coke in refrigerator.}
This task evaluates collaborative object placement in an interactive environment. One humanoid manages the refrigerator state by opening and maintaining access to the door, while the other humanoid transports the coke and places it inside. The task involves approaching, whole-body manipulation, environment interaction, object transport, and final placement.

\noindent\textbf{Bottle handoff.}
This task focuses on direct object transfer between humanoids. The receiving humanoid must coordinate with the partner's motion, wait for the appropriate transfer moment, acquire the bottle, and complete the subsequent placement task. It requires temporal synchronization, handover coordination, grasping, and transportation.

\noindent\textbf{Bottle relay.}
This task extends direct handoff into a longer relay-style collaborative sequence involving sequential object transfer and navigation. One humanoid places the bottle at an intermediate location, after which the other approaches and grasps it, transports it, and releases it at the target location. The task requires coordination across multiple stages, including partner readiness, object placement, acquisition, transportation, and final delivery.

\noindent\textbf{Bottle into cart.}
This task evaluates collaborative manipulation under changing object and environment configurations. One humanoid pushes and positions the cart, while the other approaches the bottle, grasps it, transports it toward the cart, and places it inside. The task requires coordination between cart motion and object transportation, emphasizing the coupling between locomotion, manipulation, and precise placement.

\noindent\textbf{Push hospital bed.}
This task represents a contact-rich whole-body collaboration scenario. One humanoid maintains interaction with the hospital bed while coordinating with the environment, and the other humanoid pushes the articulated object through a constrained doorway. The task requires sustained contact, locomotion coordination, forceful interaction, and long-horizon motion planning.

The environments are instantiated using RoboCasa-derived scenes~\citep{nasiriany2024robocasa} and custom-built scenarios. These environments provide diverse object configurations and interaction structures while remaining independent of the MATE collection framework.

\section{Experiments}
\label{sec:experiments}

We evaluate MATE from four perspectives. First, we assess whether the collected demonstrations support diverse policy architectures on long-horizon collaborative tasks. Second, we quantify the scalability of virtual data collection against physical teleoperation. Third, we evaluate the effectiveness of EAIS by comparing it with uniform temporal sampling. Finally, we examine zero-shot sim-to-real transfer on \textit{Bottle relay}.

\subsection{Experimental Setup}

\paragraph{Tasks and Evaluation Protocol.}
We evaluate MATE on the five long-horizon humanoid collaboration tasks introduced in the Dataset section. In each task, one humanoid is controlled by the learned policy, while the other humanoid serves as a collaborative partner following the corresponding task procedure. The partner trajectories are collected from human-operated demonstrations, providing natural variation in motion and timing.
All policies are trained on the MATE demonstrations processed by our execution-aligned interaction sampling strategy, unless otherwise specified. During evaluation, the learned policy receives only robot-centric visual observations and proprioceptive states, while the complete multi-agent interaction state is available only during data collection. This setting evaluates whether policies trained from MATE demonstrations can infer interaction progress and coordinate with other agents under partial observations. 
Each policy is evaluated with 10 full-horizon rollouts from task initialization. Grasp, transport, placement, and pushing skills are additionally evaluated with 10 stage-conditioned trials initialized from standardized stage-entry states. Grasp trials begin near the target object or handle, while subsequent skills begin from the corresponding post-grasp or post-reception states. Approach and reception are evaluated only within the full-horizon rollouts. All methods use identical demonstrations, environments, partner procedures, and evaluation budgets.

\paragraph{Policy Baselines.}
We consider five representative policy families: ACT~\citep{zhao2023aloha}, Diffusion Policy (DP)~\citep{chi2023diffusionpolicy}, $\psi_0$~\citep{wei2026psi0}, GR00T N1.7~\citep{nvidia2026grootn17}, and $\pi_{0.5}$~\citep{physicalintelligence2025pi05}. ACT and DP are trained from scratch, whereas the three vision-language-action models are fine-tuned from their official pretrained checkpoints. All policies predict a SONIC-compatible control representation used by the humanoid controller.

\begin{table*}[ht!]
\centering
\footnotesize
\setlength{\tabcolsep}{2.5pt}
\renewcommand{\arraystretch}{1.05}
\begin{tabular}{l|cccc|cc|ccc|cccc|ccc}
\toprule
&\multicolumn{4}{c}{\textbf{Coke in Refrigerator}}
&\multicolumn{2}{|c}{\textbf{Bottle Handoff}}
&\multicolumn{3}{|c}{\textbf{Bottle Relay}}
&\multicolumn{4}{|c}{\textbf{Bottle into Cart}}
&\multicolumn{3}{|c}{\textbf{Push Hospital Bed}}
\\
\cmidrule(lr){2-5}\cmidrule(lr){6-7}\cmidrule(lr){8-10}\cmidrule(lr){11-14}\cmidrule(lr){15-17}

\textbf{Baseline}
&Appr.&Grasp&Trans.&Place
&Recv.&Place
&Appr.&Grasp&Place
&Appr.&Grasp&Trans.&Place
&Appr.&Grasp&Push
\\

\midrule
$\psi_0$
&\textbf{10}|--&\textbf{9}|\textbf{10}&\textbf{6}|7&\textbf{5}|7
&6|--&\textbf{2}|2
&\textbf{10}|--&\textbf{9}|\textbf{10}&\textbf{9}|\textbf{10}
&\textbf{10}|--&\textbf{6}|\textbf{9}&\textbf{6}|\textbf{10}&\textbf{5}|\textbf{10}
&\textbf{10}|--&\textbf{7}|\textbf{7}&\textbf{7}|\textbf{10}
\\
GR00T
&7|--&4|8&4|\textbf{9}&2|\textbf{9}
&0|--&0|\textbf{4}
&5|--&4|\textbf{10}&4|9
&7|--&\textbf{6}|8&5|5&4|8
&8|--&4|6&0|5
\\
$\pi_{0.5}$
&5|--&2|3&0|5&0|0
&0|--&0|0
&1|--&0|3&0|2
&8|--&4|6&4|\textbf{10}&0|7
&5|--&1|1&0|6
\\
ACT
&0|--&0|6&0|1&0|0
&\textbf{9}|--&0|0
&0|--&0|5&0|8
&8|--&3|8&0|1&0|9
&4|--&2|5&0|7
\\
DP
&0|--&0|0&0|0&0|0
&0|--&0|0
&3|--&0|2&0|0
&0|--&0|0&0|0&0|1
&8|--&1|2&0|4
\\

\bottomrule
\end{tabular}

\caption{Long-horizon policy evaluation on MATE collaboration tasks. Each entry reports full-horizon stage completion $|$ stage-conditioned skill completion over 10 trials; ``--'' denotes not applicable.}
\label{tab:main_results}
\end{table*}
\subsection{Learning from MATE Demonstrations}
\label{sec:policy_benchmark}
Table~\ref{tab:main_results} compares full-horizon and stage-conditioned performance across policy architectures trained on MATE demonstrations.

The results show that MATE provides effective collaboration demonstrations for diverse policy architectures. $\psi_0$ achieves consistent performance across multiple tasks, successfully learning behaviors including object acquisition, transportation, and cooperative manipulation. Other policies can also acquire individual skills, but often struggle with completing full tasks from initialization, indicating the difficulty of maintaining long-horizon coordination.

The gap between full-horizon and stage-conditioned evaluation shows how temporal and interaction errors accumulate over long-horizon execution, compounding limitations in individual skills. This motivates interaction-aware temporal learning strategies such as EAIS for better utilization of long-horizon collaboration demonstrations.

\subsection{Scalability of MATE Data Collection}
\label{sec:collection_efficiency}
MATE reduces the amortized collection time per successful \textit{Bottle relay} episode from 89 seconds with physical teleoperation to 41 seconds, yielding a 2.17$\times$ improvement. Detailed timing protocols and qualitative comparisons of reset, recovery, and hardware requirements are provided in the supplementary material and accompanying videos.

\subsection{Analysis of Execution-Aligned Interaction Sampling}
\label{sec:eais_ablation}

Long-horizon teleoperation demonstrations contain redundant patterns such as hesitation, unnecessary adjustments, and task-irrelevant waiting that may be over-represented under uniform sampling. We compare EAIS with uniform sampling under identical training settings to evaluate whether prioritizing task-relevant windows improves demonstration utilization.

We train $\psi_0$ and GR00T with either uniform sampling or EAIS while keeping the demonstrations, model architectures, observation spaces, and optimization budgets unchanged. As shown in Table~\ref{tab:eais_analysis}, EAIS improves full-task completion and reduces the execution time of successful rollouts. These gains indicate that emphasizing task-progressing and interaction-critical behaviors reduces inefficient waiting and delayed responses in long-horizon partner coordination while preserving essential interaction context.

\begin{table}[!t]
\centering

\setlength{\tabcolsep}{1.5pt}
\renewcommand{\arraystretch}{1.05}
\footnotesize

\begin{tabular*}{\linewidth}{@{\extracolsep{\fill}}lcc@{}}
\toprule

\textbf{Policy}
&
\shortstack{\textbf{Coke}\\\textbf{Success (\%) / Time (s)}}
&
\shortstack{\textbf{Bottle Relay}\\\textbf{Success (\%) / Time (s)}}
\\

\midrule

$\psi_0$(Uniform)
& 10 / 157.7
& 40 / 72.8
\\

$\psi_0$(EAIS)
& 50 / 94.4
& 90 / 51.1
\\

\midrule

GR00T(Uniform)
& 0 / --
& 20 / 85.1
\\

GR00T(EAIS)
& 20 / 127.9
& 40 / 66.6
\\

\bottomrule
\end{tabular*}
\caption{EAIS versus uniform temporal sampling. Entries report full-task success (\%) / mean execution time over successful trials (s).}
\label{tab:eais_analysis}
\end{table}

\subsection{Zero-Shot Transfer of MATE Demonstrations}
\begin{figure}[!h]
\centering
\includegraphics[width=1.0\linewidth]{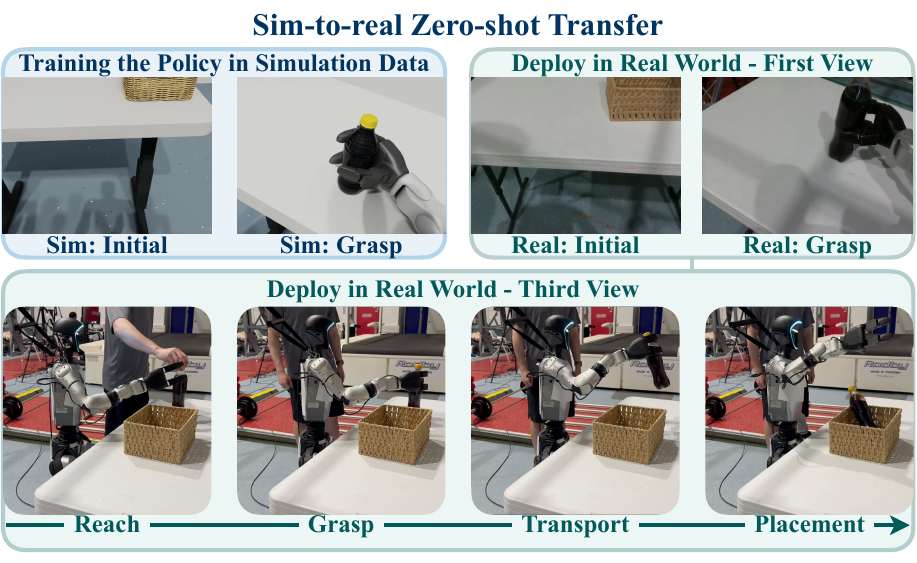}
\caption{Sim-to-real zero-shot evaluation.} 
\label{fig:sim2real}
\end{figure}

Finally, we evaluate whether MATE-collected virtual demonstrations can provide transferable supervision for real-world humanoid collaboration. We select the \textit{Bottle relay} task, which requires long-horizon behaviors including coordinating with a partner who places the bottle at an intermediate location, approaching and grasping the bottle, transporting it, and placing it at the target location.
The policy is trained solely on MATE simulation demonstrations without any real-world data or fine-tuning. During real-world evaluation, the learned policy directly controls the physical humanoid, while a human collaborator performs the partner-side stage of the task by placing the bottle at the intermediate location with natural variation in position and timing across trials. The humanoid then autonomously approaches, grasps, transports, and places the bottle using only robot-centric visual observations and proprioceptive feedback. 
Figure~\ref{fig:sim2real} contrasts the simulated training setup with a representative real-world rollout covering reaching, grasping, transportation, and placement.
Across 10 simulated and 10 real-world trials, the policy achieves 90\% success in simulation and 70\% success in real-world execution. These results demonstrate that MATE demonstrations capture transferable collaboration patterns beyond low-level visual correspondence, including coordination with the partner's object placement, visually guided object acquisition, transportation, and long-horizon task execution.

Failures mainly arise from residual sim-to-real gaps in perception, object geometry, and contact dynamics. Nevertheless, successful deployment without real-world demonstrations validates MATE as a scalable source of transferable collaboration experiences.

\section{Conclusion}

In this work, we introduced MATE, a multi-agent virtual teleoperation platform for scalable humanoid collaboration data collection. By enabling multiple operators to control humanoids in a shared physics-based environment, MATE provides a more scalable, efficient, and cost-effective paradigm for acquiring interaction-rich humanoid collaboration experiences than conventional physical collection.
Based on MATE, we demonstrated that virtual collaboration demonstrations can support long-horizon humanoid learning across diverse loco-manipulation tasks. We further introduced EAIS to better utilize these demonstrations by aligning temporal sampling with executed behaviors and interaction-critical events. Extensive evaluations across multiple learning frameworks and a zero-shot deployment on a physical humanoid show that MATE-collected demonstrations can transfer from simulation to real-world execution without additional real-world data.
MATE establishes a foundation for scalable humanoid collaboration data collection and opens new opportunities for learning complex multi-agent embodied behaviors in increasingly realistic and diverse environments.
\section*{Acknowledgments}

This work was supported by the National Natural Science Foundation of China (Project Number 62595774), MoE Key Laboratory of Intelligent Perception and Human-Machine Collaboration (KLIP-HuMaCo), HPC Platform of ShanghaiTech University.

We gratefully acknowledge Shanghai University of Sport for providing access to the experimental facilities and equipment.



\bibliography{aaai2027}
\clearpage
\appendix
\section{Supplementary Material}
\label{sec:supplementary}

This supplementary material provides implementation and evaluation details for MATE and EAIS. The material is organized as follows. \emph{Supplementary Video} summarizes the demonstration videos available on our project page. \emph{MATE Platform Details} and \emph{Dataset and Task Details} describe the multi-user system, joint episode organization, collaborative tasks, and training-set construction. \emph{Execution-Aligned Interaction Sampling} specifies the EAIS implementation, while \emph{Policy Training and Evaluation Details} and \emph{Failure Case Analysis} provide the training, evaluation, and diagnostic protocols. Finally, \emph{Scalability of MATE Data Collection} and \emph{Sim-to-Real Deployment Details} present the collection-efficiency study and physical deployment procedure.

\subsection{Supplementary Video}
\label{sec:supp_video_index}

Demonstration videos are available on our project page:
\url{https://yerik-yu.github.io/MATE/}. These videos present the MATE overview and multi-user data-collection workflow, the physical-versus-virtual collection-efficiency comparison, demonstrations of all five collaborative tasks, representative $\psi_0$ and GR00T N1.7 policy rollouts, a comparison between EAIS and uniform temporal sampling, and zero-shot sim-to-real deployment on the physical humanoid.

\subsection{MATE Platform Details}
\label{sec:supp_platform}

\paragraph{Independent embodied interfaces.}
MATE associates each remote operator with one humanoid through an independent VR interface. The VR-to-humanoid mapping is provided by the SONIC whole-body control interface used in our system. It converts operator motion into coordinated locomotion, body, arm, and hand control signals, while MATE manages multi-user synchronization, shared interaction, and joint data recording. During teleoperation, MuJoCo renders robot-specific left- and right-eye head-camera views in real time and composes them into a side-by-side stereo stream. Each stream is transmitted through the corresponding operator--robot video channel at 30 fps, providing low-latency embodied visual feedback for interactive control.

\paragraph{Multi-user isolation and shared synchronized execution.}
Each operator--humanoid pair runs in an independent worker process and is bound to a unique XR device identifier and robot index. Robot-specific ZMQ endpoints, DDS domains and topic prefixes, hand-control channels, and visual streams isolate the corresponding input and feedback paths. The XR multiplexer routes packets through robot-indexed topics, and runtime validation rejects conflicting device identifiers or communication channels before collection begins.

All active operator--robot command streams are synchronized before the shared simulation advances. The humanoids, manipulated objects, articulated furniture, and environment then evolve under the same MuJoCo physics state.

At each control step, the platform follows the sequence
\begin{enumerate}
    \item receive the latest command for every active operator--humanoid pair;
    \item align the command streams to the common simulation step;
    \item apply all commands to the shared physical world;
    \item advance contact and rigid-body dynamics;
    \item record synchronized robot, object, command, event, and camera data; and
    \item return updated robot-centric observations to the operators.
\end{enumerate}
The collection and downstream control frequency is 50~Hz.

\paragraph{Physics and rendering.}
The online and offline rendering pipelines serve different purposes. During collection, MuJoCo advances the shared physical simulation and provides low-latency stereo feedback for interactive control. After collection, the synchronized trajectories are replayed in Isaac Sim to generate standardized, higher-fidelity policy observations and supplementary visualizations. This offline re-rendering changes the camera configuration and visual appearance while preserving the recorded robot and object trajectories and their temporal alignment.

\paragraph{Joint episode recorder.}
Every execution is stored under a common simulation clock together with the episode and scene manifests. The recorder preserves generalized robot states and controller commands, base orientations and projected gravity, hand states and commands, dynamic object and articulated-joint states, contact and event streams, camera definitions, and SONIC-compatible action targets. Frame and episode indices maintain deterministic alignment among robot motion, object motion, interactions, and rendered observations.

\subsection{Dataset and Task Details}
\label{sec:supp_dataset}

\paragraph{Dataset scale and splits.}
The MATE dataset contains 2,500 joint episodes over five tasks, corresponding to 24.1 hours of synchronized collaboration data at 50~Hz. Table~\ref{tab:supp_dataset_statistics} reports the complete dataset statistics. For each task, 450 joint episodes are used for training, 25 for validation, and 25 for testing. Policy performance is evaluated separately through 10 closed-loop trials per task or stage under standardized evaluation initializations.

\begin{table}[t]
\centering
\small
\begin{tabular}{lr}
\hline
Property & Value \\
\hline
Tasks & 5 \\
Joint episodes & 2,500 \\
Episodes per task & 500 \\
Robot-centric trajectories & 5,000 \\
Synchronized time steps & 4.35M \\
Total duration & 24.1 h \\
Control / recording rate & 50 Hz \\
Train / val. / test per task & 450 / 25 / 25 \\
\hline
\end{tabular}
\caption{Summary of the MATE collaboration dataset.}
\label{tab:supp_dataset_statistics}
\end{table}

\paragraph{Episode organization.}
Each joint episode stores both humanoids under a shared temporal index, with robot-specific camera keys preserving the corresponding egocentric observations. For policy training, a model-specific adapter selects the evaluated robot while retaining the partner-conditioned scene evolution. The model-facing view consists of the selected robot's egocentric observation, proprioceptive state, task instruction, and SONIC-compatible action target. The 78-dimensional action target contains a 64-dimensional whole-body motion command and two 7-dimensional hand commands. For GR00T N1.7, we follow its official \texttt{UNITREE\_G1\_SONIC} embodiment configuration, which uses a 46-dimensional state and a 78-dimensional SONIC action. For ACT, Diffusion Policy, and $\pi_{0.5}$, the model-facing adapters use the same selected-robot 46D state and 78D action representation for consistency. The $\psi_0$ adapter follows its official 43-dimensional state interface while retaining the same 78-dimensional action target.

\paragraph{Training-set augmentation.}
The final dataset combines 903 human-operated joint episodes with 1,597 validated variants derived from them, yielding 2,500 episodes in total. For manipulation tasks, geometric augmentation introduces task-relevant perturbations to object positions at grasp stages and to handover or other interaction locations. The corresponding local motion segments are adjusted to preserve hand--object and inter-agent relationships, after which the augmented interaction is merged back into the subsequent task sequence. For Push Hospital Bed, temporal augmentation inserts additional waiting before the passage becomes available and shifts the door-opening and bed-pushing trajectories accordingly; the original motion segments are replayed at their original speed rather than stretched. Augmented variants are retained only after task-specific checks for geometric consistency, motion smoothness, contact validity, and replay stability.

\paragraph{Collaborative task roles.}
Table~\ref{tab:supp_task_roles} summarizes the two roles in each joint episode. The learned policy controls the focal humanoid listed in the third column during the reported evaluation.

\begin{table*}[t]
\centering
\small
\begin{tabular}{p{0.17\textwidth}p{0.30\textwidth}p{0.43\textwidth}}
\hline
Task & Partner role & Learned-policy humanoid role \\
\hline
Coke in Refrigerator & Approach, open, and hold the refrigerator door to maintain access. & Approach the coke, grasp it, transport it to the refrigerator, and place it inside. \\
Bottle Handoff & Present and transfer the bottle directly to the receiving humanoid. & Coordinate with the partner, receive and secure the bottle, then transport and place it. \\
Bottle Relay & Place the bottle at an intermediate location with natural variation in position and timing. & Approach the placed bottle, grasp it, transport it, and place it at the final target. \\
Bottle into Cart & Push and position the cart during the collaborative sequence. & Acquire the bottle, transport it toward the positioned cart, and place it inside. \\
Push Hospital Bed & Approach, open, and hold the corridor door clear for passage. & Approach and grip the hospital bed, wait for the passage to become available, and push the complete bed through the doorway. \\
\hline
\end{tabular}
\caption{Roles of the two humanoids in the five MATE tasks.}
\label{tab:supp_task_roles}
\end{table*}

The tasks jointly cover direct transfer, sequential relay, partner-conditioned object placement, articulated-environment interaction, and sustained contact with a large object. They require combinations of approach, grasping or contact establishment, waiting for partner progress, transport, placement, pushing, and long-horizon coordination.

\subsection{Execution-Aligned Interaction Sampling}
\label{sec:supp_eais}

The main paper defines the candidate window $W_t^i$, the execution-aligned prefix, and the final factorized weight $w_t^i$. This section specifies the implementation used in our experiments. EAIS constructs candidate demonstration windows with $H=40$ and evaluates the first $H_e=20$ steps at 50~Hz. These values define the sampling manifest and are independent of the native action chunk or runtime replanning schedule of a downstream policy.

For robot $i$, EAIS forms four state groups: base, body, hands, and non-robot scene entities. Its score uses the state of robot $i$, manipulated-object and articulated-environment states, and automatically detected contacts. The other robot's state and action signals are not used. The five signal types follow the main paper: task progress, motion-mode transitions, action-magnitude changes represented by controlled-robot activity, object and scene changes, and interaction events. No manually annotated task phases are required.

\paragraph{Automatic events and execution-prefix activity.}
Robot and scene activity are computed from normalized displacement and velocity evidence. Dataset-level activity scales use the positive 90th percentiles, while coordinate scales for target-distance evaluation are computed per episode. EAIS detects changes in the dominant robot activity group, controlled-robot and scene-motion onsets, and contact acquisition or release. Motion onsets use thresholds $0.40$ and $0.45$ for robot and scene activity, respectively, with short persistence checks and a 25-step refractory interval. Nearby events are merged, and lower-priority events immediately preceding a higher-priority event are removed.

For group $g$, the maximum displacement within the execution prefix is
\begin{equation}
\Delta_g^i(t)=
\max_{1\leq h\leq H_e}
\sqrt{
\frac{1}{|g|}
\sum_{j\in g}
\left(x_{t+h,j}^i-x_{t,j}^i\right)^2
}.
\end{equation}
With group-specific floors
$(\epsilon_{\mathrm{base}},\epsilon_{\mathrm{body}},
\epsilon_{\mathrm{hands}},\epsilon_{\mathrm{scene}})
=(0.015,0.040,0.010,0.010)$, the absolute-progress indicator is
\begin{equation}
A_t^i=
\mathbf{1}\!\left[\exists g:\Delta_g^i(t)\geq\epsilon_g\right].
\end{equation}

\paragraph{Progress and transition scoring.}
For a candidate beginning at $t$, EAIS selects the first automatic event at least five steps in the future. If no such event exists, it uses $t+80$; the target is capped at $t+100$ and the end of the episode. Let $\gamma_g^i(t)$ denote the normalized RMS displacement of group $g$ from the candidate start at $t$ to the selected target frame. Groups satisfying
$\gamma_g^i(t)\geq\max(0.25,0.30\max_{g'}\gamma_{g'}^i(t))$
form the active set $\mathcal{G}_t^i$. If this set is empty, the group with the largest $\gamma_g^i(t)$ is retained.

The execution prefix is evaluated at offsets $0,5,10,15,$ and $20$. Let $\ell_b^i(t)$ denote the average normalized distance of the active groups from the selected target at checkpoint $5b$. EAIS computes
\begin{equation}
\begin{array}{rcl}
q_t^i
&=& \displaystyle \frac{1}{4}\sum_{b=1}^{4}
\mathbf{1}\!\left[
\ell_{b-1}^i(t)-\ell_b^i(t)>-\eta_t^i
\right],\\[4pt]
n_t^i
&=& \displaystyle \mathrm{clip}\!\left(
\frac{\ell_0^i(t)-\ell_4^i(t)}
{\max(\ell_0^i(t),10^{-6})},-2,2
\right).
\end{array}
\end{equation}
where $\eta_t^i=\max(0.01,0.02\ell_0^i(t))$. Let $L_t^i$ be the longest run with controlled-robot activity below $0.22$, and let $R_t^i$ indicate activity of at least $0.25$ within the first four steps. The continuous-progress score is
\begin{equation}
\begin{array}{rcl}
s_t^i
&=& 0.45\,\mathrm{clip}\!\left(
\frac{q_t^i-0.5}{0.5},0,1\right)\\[2pt]
&& {}+0.40\,\mathrm{clip}\!\left(
\frac{n_t^i}{0.35},0,1\right)\\[2pt]
&& {}+0.15\,\mathrm{clip}\!\left(
1-\frac{L_t^i}{8},0,1\right).
\end{array}
\end{equation}
A continuous-progress window requires
$q_t^i\geq0.75$, $n_t^i\geq0.08$, $L_t^i\leq5$,
$R_t^i=1$, and $A_t^i=1$, and receives base priority
$0.20+0.80s_t^i$.

An event transition is aligned at offsets 4--16. Generic transitions use the same consistency checks with a relaxed net-progress threshold of $0.04$, whereas contact acquisition and release may instead qualify through measurable hand or scene displacement. Qualified generic transitions use $b_t^{i,\mathrm{trans}}=0.85+0.15s_t^i$, while qualified contact-acquisition and contact-release transitions use $b_t^{i,\mathrm{trans}}=0.95+0.05s_t^i$; otherwise, $b_t^{i,\mathrm{trans}}=0$. EAIS also retains a small amount of resume context: a window must begin with low activity and contain sustained controlled-robot motion resuming within the execution prefix. Such a window receives base priority $0.06$ only when it is not already classified as progress or transition.

\paragraph{Final priorities and integer sampling plan.}
Let $b_t^{i,\mathrm{cont}}$ and $b_t^{i,\mathrm{trans}}$ denote the base priorities assigned to qualified continuous-progress and event-transition windows, respectively, and let $U_t^i\in\{0,1\}$ indicate a qualified resume-context window. Using the notation of the main paper, the task-progression priority is
\begin{equation}
w_t^{i,\mathrm{prog}}=
\max\!\left(
b_t^{i,\mathrm{cont}},
b_t^{i,\mathrm{trans}},
0.06U_t^i
\right).
\end{equation}
The interaction factor is
\begin{equation}
w_t^{i,\mathrm{int}}=
\left\{
\begin{array}{ll}
8, & \mbox{contact acquisition},\\
6, & \mbox{contact release},\\
3, & \mbox{pre-contact lead},\\
1.25, & \mbox{other transition},\\
1, & \mbox{continuous progress or resume context}.
\end{array}
\right.
\end{equation}
A pre-contact lead is a certified progress window whose next contact acquisition lies beyond the aligned-transition range but within the 40-step candidate window. The unnormalized priority and nominal distribution are
\begin{equation}
w_t^i=w_t^{i,\mathrm{prog}}w_t^{i,\mathrm{int}},
\qquad
p_t^i=\frac{w_t^i}{\sum_k w_k^i}.
\end{equation}
Candidates with $w_t^i=0$ are not drawn.

The nominal priorities are converted into an exact capped integer plan using proportional allocation with largest-remainder rounding. A multiplicity $m_t^i\in\{0,\ldots,8\}$ is assigned to each start index, giving
\begin{equation}
\widetilde{p}_t^i=
\frac{m_t^i}{\sum_k m_k^i}.
\end{equation}
The allocation enforces an effective-sample-size fraction of at least $0.20$ and preserves contact-event coverage when a valid aligned window is available. Resume-context draws target $0.75\%$ of sampled starts and are capped at $1\%$. Thus, coordination-relevant waiting remains represented, while prolonged idle behavior or jitter without robot progress, scene change, or interaction events does not dominate training.

The same start-index priorities are used across the evaluated policy families, while each policy retains its native architecture, action horizon or chunk length, and runtime execution schedule.

\subsection{Policy Training and Evaluation Details}
\label{sec:supp_training_eval}

\paragraph{Training protocol.}
ACT and Diffusion Policy are trained from scratch. $\psi_0$, GR00T N1.7, and $\pi_{0.5}$ are fine-tuned from their official pretrained checkpoints. Each policy configuration is trained once, and the final designated checkpoint is used for evaluation. Table~\ref{tab:supp_training_configs} lists the verified configurations used in the experiments.

\begin{table*}[t]
\centering
\small
\begin{tabular}{p{0.09\textwidth}p{0.10\textwidth}p{0.13\textwidth}p{0.14\textwidth}p{0.31\textwidth}}
\hline
Policy & Initialization & State / action & Horizon or chunk & Optimization \\
\hline
ACT & From scratch & 46D / 78D & 100-action chunk & Batch 8; 2,000 epochs (approximately 114k optimizer steps); AdamW; fixed learning rate $10^{-5}$; weight decay $10^{-4}$. \\
Diffusion Policy & From scratch & 46D / 78D & Observation 2, prediction 16, execute 8 & Batch 16; 150k optimizer steps; AdamW with learning rate $10^{-4}$; 500-step warmup and cosine decay; 100 diffusion train and inference steps. \\
$\psi_0$ & Official pretrained checkpoint & 43D / 78D & 30-action chunk & Global batch 128; 40k optimizer steps; learning rate $10^{-4}$; 1k-step warmup and cosine decay; bf16; VLM frozen; 1,000-step flow training schedule. \\
GR00T N1.7 & Official GR00T N1.7 and Cosmos checkpoints & 46D / 78D & 40-action target & Global batch 32; 20k optimizer steps; AdamW with learning rate $10^{-4}$; state dropout probability $0.2$. \\
$\pi_{0.5}$ & Official $\pi_{0.5}$ base checkpoint & 46D / 78D & 40-action target & Global batch 4; 20k optimizer steps; AdamW; 1k-step warmup; cosine decay from $2.5\!\times\!10^{-5}$ to $2.5\!\times\!10^{-6}$. \\
\hline
\end{tabular}
\caption{Verified policy configurations used for the MATE experiments.}
\label{tab:supp_training_configs}
\end{table*}

ACT uses the original temporal aggregation mechanism with coefficient $k=0.01$. For $\pi_{0.5}$, adapting the model to the 78-dimensional action interface requires reinitializing its action-encoding and action-output projections, while retaining all shape-compatible parameters from the official base checkpoint.

\paragraph{Partner execution.}
In simulation, the collaborating humanoid follows replayed state trajectories collected from human-operated MATE demonstrations. Different partner trajectories are used across rollouts, producing variation in motion, interaction timing, object placement, and articulated-environment state while preserving the corresponding task procedure. The evaluated policy receives only its robot-centric visual observation and proprioceptive state.

\paragraph{Full-horizon and stage-conditioned evaluation.}
Every reported result uses the formal 50-Hz closed-loop evaluation pipeline. Each full-horizon entry is measured over 10 trials beginning from task initialization. Each applicable stage-conditioned skill is measured over 10 trials initialized from standardized stage-entry states. Approach and reception are evaluated as parts of full-horizon execution, because their relevant behavior begins at the start of the episode rather than from an isolated near-interaction initialization.

A rollout succeeds only when the task- or stage-specific completion condition is reached. A timeout, robot fall, dropped object, loss of required sustained contact, or failure to satisfy the completion criterion is counted as failure. Execution time in the EAIS analysis is averaged over successful trials.

\paragraph{Uniform-sampling ablation.}
The uniform ablation uses the same demonstrations, train/validation/test split, eligible start-index support, boundary masks, model architecture, state and action representation, optimization budget, evaluation initializations, and partner procedures as the corresponding EAIS run. Only the probability assigned to each eligible start index changes: EAIS uses $\widetilde{p}_t^i$, whereas the uniform ablation assigns equal probability to all eligible starts. Zero-weight terminal or otherwise ineligible regions remain excluded in both settings.

\subsection{Failure Case Analysis}
\label{sec:supp_failure}

The difference between full-horizon and stage-conditioned results reveals how interaction and skill errors accumulate across collaborative execution. The stage-conditioned results in Table~\ref{tab:main_results} show that ACT and Diffusion Policy can acquire local grasping behaviors when the relevant object and interaction state are already nearby. Full-horizon execution additionally requires the policy to approach the correct region, respond to partner progress, preserve object and contact state, and transition between skills without an externally supplied stage boundary.

\paragraph{Temporal coordination.}
Long-horizon tasks contain periods in which the focal humanoid must continue approaching, wait for a partner-conditioned state change, or act promptly after the interaction becomes available. Inefficient waiting delays later stages, whereas an early response can lead to an unfavorable approach or contact configuration. These timing errors compound because the resulting state becomes the starting condition for every subsequent behavior. Representative failures exhibit this long-horizon sensitivity: delayed responses or inefficient waiting alter the state from which later approach, grasp, and placement behaviors must proceed.

\paragraph{Approach and target localization.}
Small errors in base orientation, approach distance, or visual target localization can leave the hands outside the reliable grasp region. A policy may therefore exhibit plausible navigation while entering manipulation from a state that differs from the demonstration distribution. Stage-conditioned evaluation reduces this source of variation by initializing the robot near the relevant object or handle, explaining why local skill completion can be substantially higher than full-task completion.

\paragraph{Contact establishment and object retention.}
Bottle Handoff and Bottle Relay require contact to be established at the correct object pose and interaction time. A marginal initial grasp can become unstable during lift or transport, causing slip, object loss, or an unfavorable arm configuration. Similarly, Push Hospital Bed requires sustained two-hand interaction with a large articulated object. Loss of contact or asymmetric pushing can redirect the bed and prevent passage through the doorway.

\paragraph{Skill transitions.}
A successful grasp does not by itself guarantee successful transport or placement. The policy must preserve the object while changing whole-body motion, navigate toward the target, and execute a controlled release. The representative policy rollouts in the supplementary demo video include successful full-horizon examples in which GR00T or $\psi_0$ maintains the required interaction state across multiple stages.

\paragraph{Effect of EAIS.}
Uniform temporal sampling assigns a large portion of training exposure to frequently occurring behavior. EAIS increases the representation of progress, contact transitions, and prompt responses after partner-conditioned state changes while retaining a small amount of meaningful waiting context. The improvements in full-task success and successful-rollout duration reported in the main paper are consistent with fewer delayed responses and more reliable transitions across long-horizon coordination.

\subsection{Scalability of MATE Data Collection}
\label{sec:supp_scalability}

We compare MATE with physical teleoperation on the same Bottle Relay task procedure. Physical collection time is measured from task start to task completion, whereas the reported MATE time is amortized over successful episodes and includes virtual environment reset. Failed retries are excluded from both methods. Physical collection requires 89 seconds per successful episode on average, whereas MATE requires 41 seconds, corresponding to a $2.17\times$ speedup in successful-episode collection.

Beyond the measured task-execution time, MATE changes the infrastructure required to scale collaborative collection. Physical collection couples the number of simultaneously controlled agents to the availability of complete humanoid systems, a sufficiently large and safe workspace, synchronized on-site operation, and physical initialization and recovery. MATE instead instantiates multiple humanoids in a shared virtual environment and permits geographically distributed operators to enter the same episode through independent interfaces. New tasks can reuse the multi-user collection system while changing the scene, objects, and role specification. The supplementary demo video presents the collection workflow and the qualitative contrast between the physical and virtual pipelines.

\subsection{Sim-to-Real Deployment Details}
\label{sec:supp_sim2real}

For sim-to-real training, the recorded MATE Bottle Relay trajectories are first replayed in a deployment-aligned virtual scene. The scene layout, camera configuration, and visual appearance are adapted to approximate the physical setup while preserving the recorded robot and object trajectories, task structure, and interaction timing. The resulting robot-centric observations are used to train the $\psi_0$ policy.

The policy is deployed through the official $\psi_0$--SONIC pathway: robot-centric images and proprioceptive state are processed by $\psi_0$, and the predicted 78-dimensional actions are executed by the SONIC humanoid controller. No real-world demonstrations or real-world fine-tuning are used.

The simulated partner is another humanoid following the Bottle Relay role. In the physical evaluation, a human partner performs the same partner-side stage by placing the bottle at an intermediate location. The partner varies the bottle position and placement timing across trials. After placement, the humanoid autonomously approaches the bottle, grasps it, transports it, and places it at the target using only robot-centric visual observations and proprioceptive feedback.

We conduct 10 simulated and 10 real-world trials under this procedure. The policy achieves $90\%$ success in simulation and $70\%$ success on the physical humanoid. The demo video shows the simulated training task and representative real-world execution.


\end{document}